# LittiChoQA: Literary Texts in Indic Languages Chosen for Question Answering

**Aarya Khandelwal[1], Ritwik Mishra[2], Rajiv Ratn Shah[1]**
[1] Indraprastha Institute of Information Technology, Delhi
`aarya22007@iiitd.ac.in, rajivratn@iiitd.ac.in`
[2] Accenture, India
`ritwik.a.mishra@accenture.com`

## Abstract

Long-context question answering (QA) over literary texts poses significant challenges for modern large language models, particularly in low-resource languages. We address the scarcity of long-context QA resources for Indic languages by introducing LittiChoQA, the largest literary QA dataset to date covering many languages spoken in the Gangetic plains of India. The dataset comprises over 270K automatically generated question–answer pairs with a balanced distribution of factoid and non-factoid questions, generated from naturally authored literary texts collected from the open web. We evaluate multiple multilingual LLMs on non-factoid, abstractive QA, under both full-context and context-shortened settings. Results demonstrate a clear trade-off between performance and efficiency: full-context fine-tuning yields the highest token-level and semantic-level scores, while context shortening substantially improves throughput. Among the evaluated models, Krutrim-2 achieves the strongest performance, obtaining a semantic score of 76.1 with full context. While, in shortened context settings it scores 74.9 with answer paragraph selection and 71.4 with vector-based retrieval. Qualitative evaluations further corroborate these findings.[1]

## 1 Introduction

Literary texts often involve extended narrative passages that demand holistic comprehension for downstream tasks such as question answering (QA). In particular, QA over extended contexts introduces challenges beyond those encountered in conventional short-context settings. Modern large language models (LLMs) exhibit degraded performance as context length increases, attributable in part to limitations in self-attention mechanisms and model reasoning capacity over long spans of text (Liu et al., 2024). Furthermore, extended contexts enable the formulation of questions that demand elaborate, descriptive responses (non-factoid questions), in contrast to factoid questions, which can be addressed through the extraction of short, verbatim spans from the source text.

The scarcity of publicly available resources for long-context QA on literary texts in Indic languages constitutes the primary motivation for this study. The main contributions of our work are listed as follows: (a) we introduce the largest literary QA dataset to date, LittiChoQA (**Lit**erary **T**exts in **I**ndic Languages **Cho**sen for **Q**uestion **A**nswering)[2], comprising over 270K question–answer pairs with an approximately equal distribution of factoid and non-factoid questions; and (b) we release fine-tuned adapter checkpoints of the best-performing LLM for both long-context and truncated-context settings.

To the best of our knowledge, this is the first work to release a question-answering resource for low-resource Indic languages spoken in the Gangetic plains of India. We anticipate that this dataset will facilitate the development and evaluation of models for long-context question answering in Indic languages. Further details on the literary sources and covered languages are provided in Section 3.

## 2 Related Work

Several prominent studies on question answering (QA) over literary texts have focused on short stories drawn from school textbooks (Richardson et al., 2013; Hill et al., 2016), fables (Zhao et al., 2023), and books sourced from Project Gutenberg[3] (Kočiský et al., 2018; Xu et al., 2022), primarily in the English language. The size of these datasets typically ranges from a few hundred to fewer than a couple of thousand stories. Xu et al. (2022) demonstrated the effectiveness of employing generative models for question generation. Similarly, Zhao et al. (2023) emphasized that generative approaches can facilitate the development of QA resources and further showed that non-factoid abstractive QA poses a substantially more challenging problem.

---
[1] We release the code and resources publicly on `https://github.com/ritwikmishra/LittiChoQA/`

[2] The name is inspired by a popular dish (*Litti Chokha*) commonly eaten in the Gangetic plains of India.
[3] `https://www.gutenberg.org/`

|            | Urdu   | Hindi  | Angika | Bhojpuri | Awadhi | Maithili | Telugu  | Konkani | Bundeli |
|------------|--------|--------|--------|----------|--------|----------|---------|---------|---------|
| # Stories  | 1742   | 1157   | 196    | 133      | 124    | 101      | 97      | 61      | 58      |
| # Avg Words| 3477.3 | 3849.3 | 577.1  | 718.2    | 990.4  | 2072     | 22431.7 | 2215.7  | 578     |
| # QA       | 81566  | 56078  | 3445   | 2484     | 3119   | 4640     | 115980  | 2885    | 862     |

|            | Nepali | Braj  | Dogri | Pali  | Dzongkha | Magahi | Bagheli | Assamese | **Total** |
|------------|--------|-------|-------|-------|----------|--------|---------|----------|-----------|
| # Stories  | 32     | 32    | 30    | 27    | 18       | 8      | 6       | 2        | 3824      |
| # Avg Words| 546.3  | 951.6 | 417.9 | 436.6 | 543.2    | 1876   | 2236.7  | 1145.5   | 2650.8    |
| # QA       | 580    | 722   | 419   | 261   | 373      | 339    | 282     | 73       | 274108    |

Table 1: Distribution of languages in the LittiChoQA dataset. The table reports, for each language, the number of stories, the average number of words per story, and the number of question–answer pairs. Word counts are computed using whitespace tokenization.

A prevalent approach for constructing QA resources in Indic languages has been the machine translation of existing English QA datasets (Sabane et al., 2023; Ghatage et al., 2023; Endait et al., 2025; Singh et al., 2025). However, beyond issues related to answer alignment, this translation-based strategy is not sustainable for extremely low-resource languages, for which reliable machine translation systems are often unavailable. Anuranjana et al. (2019) introduced a reading comprehension dataset for Hindi that was compiled from Hindi stories available on the open web, and observed that factoid questions can often be addressed through simple lexical matching. Aurpa et al. (2025) proposed a Bengali QA resource in which the contexts were derived from authentic long-form Bengali texts from literature and other domains accessible on the open web.

Therefore, these observations suggest that, rather than relying on machine translation, the collection of naturally authored literary texts from the open web, combined with generative techniques, constitutes a promising direction for the development of QA resources in the literary domain.

## 3 LittiChoQA Dataset

We collected literary texts in multiple Indic languages from openly accessible web sources, ranging from public archives to copyrighted materials accessed with due permission. The data is sourced from open data repositories like Kaggle and GitHub, as well as digital libraries containing folk literature, books and novels. Scanned manuscripts were processed using Optical Character Recognition (OCR) tools to extract text in the native scripts. The list of resources is given in Appendix D. We used the GPT-4.1 model (gpt-4.1-2025-04-14) via OpenAI API calls to generate both factoid and non-factoid question–answer pairs from each story. An example of a generated non-factoid question is presented in Table 2.

| Field | Content |
|-------|---------|
| Description | The story shows the friendship of two oxen (Heera and Moti) who suffer hardships but try to return to their home safely. |
| Question | कहानी के अंत में हीरा और मोती की घर वापसी और गाँव वालों की प्रतिक्रिया से क्या संदेश मिलता है? <br> *What message is conveyed by Heera and Moti's return and the villagers' response?* |

Table 2: GPT-4.1 generated non-factoid question for short story "Do Bail" by Munshi Premchand.

The prompt employed for this process is provided in Appendix A. Due to the model's context limitation of 10K tokens per minute (TPM) under the tier-1 plan, long stories were segmented into non-overlapping chunks. Non-overlapping segmentation was chosen to avoid the generation of redundant question–answer pairs that could arise from overlapping contexts. In addition, we release chunk-level metadata for each story to enable future researchers to trace each QA pair back to the specific source segment from which it was derived. The distribution of languages in the resulting dataset is presented in Table 1.

To assess the quality of the generated dataset, we manually annotated a subset of the data, containing

214 QA pairs (factoid + non factoid). Each annotation instance consists of a story segment paired with a corresponding QA pair, which was evaluated across multiple aspects. Several of these aspects are informed by prior work, including: (a) QA readability (Yao et al., 2022), (b) question relevance and correctness (Moon et al., 2024), and (c) question difficulty (Mulla and Gharpure, 2023). In addition, we introduce two further annotation aspects: (d) answer relevance and correctness, and (e) Non-factoidness. The Likert-scale definitions used for all annotation aspects are provided in Appendix B. Annotators participated on a voluntary basis.

The annotation results indicate that the QA pairs generated using GPT-4.1 are grammatically well-formed, readable, and contextually relevant to the source narratives, while also being reasonably easy to answer. Furthermore, the annotations confirm that questions labeled as factoid and non-factoid adhere well to their intended definitions.

|  |  | (a) | (b) | (c) | (d) | (e) |
|---|---|---|---|---|---|---|
| Hindi | F (66) | 4.9 | 4.7 | 4.7 | 4.7 | 1.2 |
|  | Nf (68) | 5.0 | 5.0 | 3.7 | 5.0 | 4.0 |
| Bhojpuri | F (32) | 4.7 | 4.7 | 4.0 | 4.6 | 1.7 |
|  | Nf (29) | 5.0 | 5.0 | 4.0 | 5.0 | 3.2 |
| Awadhi | F (9) | 5.0 | 5.0 | 4.0 | 5.0 | 1.1 |
|  | Nf (10) | 5.0 | 5.0 | 4.0 | 5.0 | 3.1 |
| Total | F (107) | 4.7 | 4.8 | 4.4 | 4.7 | 1.3 |
|  | Nf (107) | 5.0 | 5.0 | 3.8 | 5.0 | 3.7 |

Table 3: Results of manual annotation across five evaluation aspects: (a) question–answer readability, (b) question relevance and correctness, (c) question difficulty, (d) answer relevance and correctness, and (e) non-factoidness. Column headers correspond to the aspect indices. F denotes factoid questions and Nf denotes non-factoid questions. Values in parentheses indicate the corresponding counts.

## 4 QA Models

We fine-tuned large language models (LLMs) on non-factoid question–answer pairs from the proposed dataset. This choice was motivated by the availability of numerous multilingual extractive QA models for factoid questions, which are typically fine-tuned on pretrained transformer encoders. In contrast, non-factoid questions with abstractive answers pose a greater level of difficulty and remain less explored (Weissenborn et al., 2017).

For this study, we selected six instruction-tuned multilingual LLMs whose pretraining data included Indic languages. The selection criteria and the complete list of models are detailed in Appendix C. The models considered were: (i) Llama 3.1 8B (Grattafiori et al., 2024), (ii) Aya 23 8B (Aryabumi et al., 2024), (iii) Sarvam-1 (Sarvam, 2022), (iv) OpenHathi-7B (Sarvam, 2023), (v) Krutrim-2 12B (Kallappa et al., 2025), and (vi) Qwen2.5 7B (Team, 2024). In addition, we evaluated context-shortening strategies as proposed by Mishra et al. (2025a).

### 4.1 Implementation

Fine-tuning was conducted using the PEFT library (Mangrulkar et al., 2022) in conjunction with low-rank adaptation (LoRA) techniques (Hu et al., 2022). The dataset was partitioned into training, development, and test sets in a 70:10:20 ratio. The hyperparameter settings are referenced in Table 5. All experiments were carried out using two NVIDIA A100 GPUs. To shorten the context conditioned on a given question, two complementary strategies were employed: (i) a fine-tuned Answer Paragraph Selection (APS) model proposed by Mishra et al. (2025b), and (ii) a vector-store–based retriever leveraging `paraphrase-multilingual-MiniLM-L12-v2` sentence embeddings (Reimers and Gurevych, 2019).

## 5 Results

Abstractive question answering models are commonly evaluated using lexical metrics such as ROUGE (Zhao et al., 2023). In this work, we adopt the evaluation metrics used by (Mishra et al., 2025a), which incorporates both lexical and semantic metrics. Table 4 reports the performance of the fine-tuned LLMs on a common test set. As expected, fine-tuning with the full context consistently outperforms fine-tuning with shortened contexts. Nevertheless, context shortening substantially improves model throughput. Throughput is defined as the number of questions the model can generate answers for within the fixed computational budget. Specifically, QA models operating on shortened contexts were able to generate answers for more than 20K questions, whereas models processing long contexts generated answers for only $\sim 900$ questions. Furthermore, context reduction using the fine-tuned APS model yielded su-perior performance compared to the vector-based retriever. Among the evaluated LLMs, Krutrim-2 demonstrated the strongest overall performance. Notably, these trends were also observed when the LLMs were evaluated without fine-tuning.

| Models | Full Context | | | | Shortened Context using ... | | | | | | | |
|---|---|---|---|---|---|---|---|---|---|---|---|---|
| | | | | | Fintuned APS | | | | Vector Retriever | | | |
| | R1 | R2 | RL | STS MuTe ↓ | R1 | R2 | RL | STS MuTe | R1 | R2 | RL | STS MuTe |
| OpenHathi | 14.1 | 4.0 | 10.2 | 52.0 | 14.2 | 4.3 | 10.2 | 48.7 | 9.7 | 2.6 | 7.3 | 43.9 |
| Sarvam-1 | 14.5 | 3.3 | 9.4 | 58.8 | 13.4 | 3.6 | 8.7 | 58.4 | 7.4 | 1.5 | 5.6 | 49.2 |
| Qwen2.5 | 11.9 | 3.2 | 8.4 | 61.9 | 11.4 | 3.4 | 8.0 | 60.4 | 6.8 | 1.8 | 5.0 | 55.9 |
| Aya 23 | 27.3 | 8.0 | 18.5 | 67.0 | 25.5 | 7.5 | 17.6 | 65.2 | 17.7 | 5.3 | 12.7 | 58.0 |
| Llama 3.1 | 36.2 | 13.5 | 24.4 | 74.7 | **35.3** | 12.6 | 23.9 | 73.9 | 29.4 | 9.8 | 20.6 | 70.8 |
| Krutrim-2 | **37.8** | **16.7** | **27.1** | **76.1** | 35.2 | **15.2** | **25.2** | **74.9** | **30.5** | **12.2** | **21.8** | **71.4** |

Table 4: Results of fine-tuned large language models evaluated on identical test samples. Performance is assessed using ROUGE metrics (R1, R2, and RL) to measure token-level similarity and STS MuTe to evaluate semantic-level similarity. The results indicate that Krutrim consistently achieves the best performance across different settings. Moreover, fine-tuning with the full context outperforms context shortening approaches, and among the context shortening methods, the APS-based model yields better results than the vector retriever.

| Hyperparameter | Value |
|---|---|
| Optimizer | AdamW |
| Learning rate | $1 \times 10^{-6}$ |
| Epochs | 1 |
| LoRA rank ($r$) and scaling factor ($\alpha$) | 32 |
| Quantization | 4-bit NF4 |
| Training precision | bf16 |

Table 5: Hyperparameter settings used for fine-tuning.

### 5.1 Qualitative Evaluation

To further analyze the performance of Krutrim-2, we conducted a qualitative evaluation in which human annotators were asked to rank model-generated answers alongside the reference answer. For three sampled instances with full context, Krutrim-2 was consistently ranked as producing the best answer after the reference. Under shortened-context settings, annotators ranked the model outputs based on their similarity to the reference answer, and Krutrim-2 was consistently placed among the top three most similar responses.

We additionally performed a comparable qualitative evaluation using GPT-4.1 as an automatic annotator in place of human judgment. The observations aligned with those obtained from human annotation on the same three examples. Consequently, GPT-4.1 was employed as a surrogate annotator for additional test samples. The results indicated that, when full context was provided, the reference answer was consistently ranked first, followed by Krutrim-2 as the second-best response.

In the shortened-context setting, when model-generated answers were compared against the reference, GPT-4.1 ranked Krutrim-2 first in 12 out of 17 languages.

## 6 Conclusion

We present LittiChoQA, a large-scale literary question answering dataset designed to address the lack of long-context QA resources for Indic languages. The dataset demonstrates that combining naturally authored literary texts with generative models is an effective alternative to translation-based data creation. Empirical analyses confirm the quality and consistency of the generated QA pairs, particularly for challenging non-factoid questions. Experimental results reveal a trade-off between answer quality and computational efficiency: full-context training delivers better performance, while context-shortening methods enable substantially higher throughput. Across all settings, Krutrim-2 emerges as the most reliable model, and APS model proves more effective than vector-based retrieval for context reduction.

### 6.1 Future Work

We aim to extend LittiChoQA to additional Indic languages, literary genres, and historical periods to improve linguistic and cultural coverage. Research on integrated long-context models with dynamic context filtering could mitigate "lost-in-the-middle" issues by focusing on relevant spans within long texts (Deng et al., 2024). Moreover, evaluating newer long-context LLMs and investigating multilingual transfer and cross-lingual generalization remain promising directions for advancing literary question answering in low-resource settings.

## 7 Ethical Considerations

All data collection was conducted in compliance with applicable usage policies and technical constraints. A single device was used for data acquisition, adhering strictly to prescribed rate limits to avoid undue load on hosting services. The literary texts were sourced from publicly accessible repositories and platforms; where copyright restrictions applied, we proactively contacted data hosting services or copyright holders. The dataset is released only after obtaining explicit consent and is intended exclusively for non-commercial research use.

In addition, care was taken to preserve the integrity of the original texts and to avoid introducing harmful, misleading, or sensitive content during processing and question–answer generation. The dataset does not contain personal or private information, and no attempt was made to infer or annotate sensitive attributes. These measures were adopted to ensure responsible data use and to support ethical research practices in low-resource language settings.

## 8 Limitations

The scanned manuscripts were processed using OCR tools whose accuracy in the target languages is not explicitly known. This may introduce transcription errors in the extracted text. Also we have only considered prose texts and not poetry related works as they feature complex structural elements such as rhyme and unconventional syntax which pose significant challenges for LLMs. The manual annotation process involved a single annotator per language, primarily due to monetary and logistical constraints, thereby done on a voluntary basis. The qualitative evaluation was conducted on a relatively small number of examples. This is attributable to the substantial time and cognitive effort required to read long literary passages, interpret questions, and compare multiple model-generated answers, which constrained the scale of human evaluation. The question–answer pairs were generated from non-overlapping chunks of long stories. Although this design choice avoids redundancy, it precludes multi-hop or cross-chunk reasoning, thereby limiting the dataset's capacity to evaluate deeper narrative understanding. We have currently reported the results on a single run focusing primarily on non-factoid, abstractive question answering. While motivated by its relative difficulty and underexploration, factoid QA over long contexts remains an important direction that is not addressed here and is deferred to future work.

# A QA Generation Prompt

The following prompt was used to generate factoid and non-factoid QA pairs from the story given to GPT-4.1 model.

```
You are given a story below in {LANGUAGE} language. Your task is to read it carefully
    and generate as many non-factoid -questionanswer (QA) pairs as possible based on
    the content of the story. Question and answer should be in the same {LANGUAGE}
    language and its script.

Each question must:
- Be non-factoid in nature (i.e., not answerable with a single word, number, or
    phrase).
- Require a long, descriptive answer that demonstrates reasoning, interpretation, or
    inference.
- Encourage answers that span multiple sentences, possibly combining information from
    various parts of the story.
- Not be limited to extractive answers; answers can be abstractive, paraphrased, or
    synthesized based on the content.

Each answer must:
- Be coherent, contextually appropriate, and logically grounded in the story.
- Demonstrate critical thinking and deep understanding of the narrative.
- Include inferences, motivations, or implications if relevant.

Use the following format for each QA pair:

##[Nf]Question N:
<Your question>
##[Nf]Answer N:
<Your answer>

When no further non-factoid questions can be generated, end the output with
    "##END[Nf]".
Then generate as many factoid -questionanswer (QA) pairs as possible based on the
    content of the story. Question and answer should be in the same language and
    script as the story. Each question must:
- Be factoid in nature (i.e., answerable with a single word, number, or phrase).
- Be directly answerable from the story without requiring inference or interpretation.
- Be clear and unambiguous.

Each answer must:
- Be a short span of text taken verbatim from the story that answers the question.

Use the following format for each QA pair:
##[F]Question N:
<Your question>
##[F]Answer N:
<Your answer>

When no further factoid questions can be generated, end the output with "##END[F]".

IMPORTANT: The generated questions and answers should be in {LANGUAGE} language and
    its script.

## BEGINNING OF STORY ##
```

## B Likert Scales For Annotation

A given story is passed to GPT-4.1 model and it was asked to generate factoid and non-factoid questions-answer pairs from the story. In order to evaluate the quality of generated QA pairs, The annotations were performed on the following five different aspects using Likert scales.

(a) QA readability

- ☐ 1: The above question and answer are completely grammatically incorrect and unreadable.
- ☐ 2: The above question and answer are somewhat grammatically incorrect and unreadable.
- ☐ 3: The above question and answer are somewhat grammatically correct but unreadable.
- ☐ 4: The above question and answer are grammatically correct and somewhat unreadable.
- ☐ 5: The above question and answer are completely grammatically correct and readable.

(b) Question relevancy

- ☐ 1: The above question is completely irrelevant to the story.
- ☐ 2: The above question is somewhat irrelevant to the story.
- ☐ 3: The above question is relevant to the story, but factually incorrect.
- ☐ 4: The above question is somewhat relevant to the story, and factually correct.
- ☐ 5: The above question is completely relevant to the story, and factually correct.

(c) Question difficulty

- ☐ 1: The above question is very difficult to answer, and requires deep understanding of the story.
- ☐ 2: The above question is difficult to answer, and requires a good understanding of the story.
- ☐ 3: The above question is somewhat difficult to answer, and requires some understanding of the story.
- ☐ 4: The above question is somewhat easy to answer, and requires some understanding of the story.
- ☐ 5: The above question is very easy to answer, and requires no understanding of the story.

(d) Answer relevancy and correctness

- ☐ 1: The above answer is completely irrelevant to the question and incorrect.
- ☐ 2: The above answer is somewhat irrelevant to the question and incorrect.
- ☐ 3: The above answer is somewhat relevant to the question but incorrect.
- ☐ 4: The above answer is somewhat relevant to the question and correct.
- ☐ 5: The above answer is completely relevant to the question and correct.

(e) Non-factoidness

- ☐ 1: The above question can be answered with a single short text (a phrase) extracted verbatim from the story.
- ☐ 2: The above question can be answered with a single long text (a sentence or paragraph) extracted verbatim from the story.
- ☐ 3: The above question requires minor abstraction or inference beyond a direct extract, but the answer is still mostly present in the story.
- ☐ 4: The above question requires combining information from multiple parts of the story, with significant inference or synthesis.
- ☐ 5: The above question cannot be answered by extracting or paraphrasing text from the story; it requires reasoning, explanation, or external/common knowledge.

## C LLM Selection Criteria

We searched the LLMs to be finetuned on Huggingface model hub[4]. The following criteria were used to select the LLMs:

- Task: Text Generation
- Frameworks: PyTorch, Transformers
- Languages: Hindi, Tamil, Urdu, Telugu
- Parameter Count: $> 1B$
- Architecture: CausalLM

### C.1 Selected Models

| # | Model Name | Model ID |
|---|---|---|
| 1 | LLaMA 3.1 8B Instruct | meta-llama/Llama-3.1-8B-Instruct |
| 2 | Aya 23 8B | CohereLabs/aya-23-8B |
| 3 | Sarvam-1 | sarvamai/sarvam-1 |
| 4 | OpenHathi-7B-Hi-v0.1-Base | sarvamai/OpenHathi-7B-Hi-v0.1-Base |
| 5 | Krutrim-2-Instruct | krutrim-ai-labs/Krutrim-2-instruct |
| 6 | Qwen2.5 7B Instruct | Qwen/Qwen2.5-7B-Instruct |

Table 6: List of selected LLMs and their corresponding identifiers

## D Language Resources

We list the URLs and digital library resources used for each of the 17 Indic languages in the dataset.

---
[4]https://huggingface.co/models

| Language | Links |
|---|---|
| Hindi | https://www.kaggle.com/datasets/amankhandelia/premchand-corpus |
| | https://github.com/midas-research/bhaav |
| | https://github.com/midas-research/hindi-discourse |
| | https://typeinhindi.com/online-hindi-ocr |
| Telugu | https://github.com/AnushaMotamarri/Telugu-Books-Dataset?tab=readme-ov-file |
| Bagheli | https://archive.org/details/dli.language.1292/ |
| Bundeli | https://archive.org/details/dli.language.1293/ |
| Awadhi | https://ia801502.us.archive.org/11/items/in.ernet.dli.2015.464097/2015.464097.Lok-Katha_text.pdf |
| Assamese | https://archive.org/details/dli.nbt.110/mode/2up |
| | https://archive.org/details/dli.nbt.112/mode/2up |
| Dogri | https://archive.org/details/dli.language.0931/ |
| Bhojpuri | https://bhojpurisahityangan.com/wp-content/uploads/2025/05/Lalpurja.pdf |
| | https://bhojpurisahityangan.com/wp-content/uploads/2025/04/Girmitya-Bharatvanshi.pdf |
| | https://bhojpurisahityangan.com/wp-content/uploads/2024/12/Johar-Bhojpuriya-Maati-Bhag-02.pdf |
| | https://bhojpurisahityangan.com/wp-content/uploads/2024/10/Johar-Bhojpuriya-Maati-_-Bhag-03.pdf |
| | https://bhojpurisahityangan.com/किताब-घर/page/5/ |
| Braj | https://ia601402.us.archive.org/34/items/in.ernet.dli.2015.379002/2015.379002.Braj-ki.pdf |
| Magahi | https://magahi-sahitya.blogspot.com/ |
| | https://magadh-ki-lok-kathayen.blogspot.com/ |
| Angika | https://kahani.angika.com/ |
| Maithili | https://khattarkaka.com/ |
| | https://maithilisamachar.com/maithili-stories/ |
| Pali | https://archive.org/details/in.ernet.dli.2015.404822/page/n9/mode/2up |
| Nepali | https://friendspeaceteams.org/nepali-story-collection/ |
| Konkani | https://archive.org/details/gerra0000shee/page/2/mode/2up |
| | https://archive.org/details/rudra0000gaja/page/n19/mode/2up |
| | https://archive.org/details/konkanikathasang0000chan/page/n35/mode/2up |
| Dzongkha | https://archive.org/details/bdrc-W00KG03678 |
| | https://www.easyocrconverter.com/free-ocr-dzongkha-language/ |
| | https://www.i2ocr.com/free-online-dzongkha-ocr |
| Urdu | https://www.rekhta.org/ |

Table 7: Language-wise datasets and resources